\documentclass{article}

\PassOptionsToPackage{numbers, compress}{natbib}
\usepackage[preprint]{neurips_2025}

\usepackage[utf8]{inputenc} % allow utf-8 input
\usepackage[T1]{fontenc}    % use 8-bit T1 fonts
\usepackage[english]{babel} % better handling of accents
\usepackage{url}            % simple URL typesetting
\usepackage{booktabs}       % professional-quality tables
\usepackage{amsfonts}       % blackboard math symbols
\usepackage{nicefrac}       % compact symbols for 1/2, etc.
\usepackage{microtype}      % microtypography
\usepackage{xcolor}         % colors

\usepackage{capt-of}
\usepackage{xspace}
\usepackage{xcolor}
\usepackage{enumitem}
\usepackage{amsmath,amssymb,amsbsy,amsfonts,dsfont,pifont,bm,bbm,mathrsfs,mathtools,nicefrac}
\usepackage{algorithm,algpseudocode,listings}
\usepackage{booktabs,multirow,adjustbox,diagbox,threeparttable,tabularray}

\definecolor{citeblue}{rgb}{0.21,0.49,0.74}
\usepackage[pagebackref,breaklinks,colorlinks,citecolor=citeblue]{hyperref}
\usepackage{subfigure}
\usepackage{wrapfig}
\usepackage[capitalize]{cleveref}  % Should be loaded after 'hyperref', and works perfectly with 'subfigure'.
\crefname{section}{Sec.}{Secs.}
\Crefname{section}{Section}{Sections}
\crefname{table}{Tab.}{Tabs.}
\Crefname{table}{Table}{Tables}
\crefname{figure}{Fig.}{Figs.}
\Crefname{figure}{Figure}{Figures}
\crefname{equation}{Eq.}{Eqs.}
\Crefname{equation}{Equation}{Equations}
\usepackage{tcolorbox}
\usepackage[utf8]{inputenc} % Ensure UTF-8 encoding
\usepackage{amsmath, amssymb, amsthm}
\usepackage{svg}
\usepackage{colortbl}
\usepackage{tabularx}
\usepackage{wrapfig}
\usepackage{microtype}

\newcommand{\tocite}[1]{{\color{red} [TO CITE]}}
\newcommand{\methodname}{MiCo}
\newcommand{\method}{\texttt{\methodname}\xspace}

\usepackage{colortbl,xcolor}

\definecolor{rliableblue}{HTML}{77AADD}
\definecolor{darkblue}{rgb}{0.0, 0.0, 0.55}
\definecolor{bblue}{rgb}{0,0.45,0.74}
\definecolor{myred}{rgb}{0.85,0.33,0.1}
\definecolor{deepgreen}{RGB}{0, 150, 0}
\definecolor{website_color}{rgb}{0.9333333333333333, 0.10980392156862745, 0.592156862745098}

\title{\method: Multi-image Contrast for Reinforcement \\ Visual Reasoning}

\author{
    Xi Chen$^{1}$ \quad
    Mingkang Zhu$^{3}$ \quad
    Shaoteng Liu$^{3}$ \quad
    Xiaoyang Wu$^{1}$ \quad
    Xiaogang Xu$^{3}$ \quad \\
    \bf
    Yu Liu$^{2}$ \quad
    Xiang Bai$^{4}$ \quad
    Hengshuang Zhao$^{1}$\\[5pt] 
    $^{1}$HKU \quad
    $^{2}$ Tongyi Lab, Alibaba Group \quad
    $^{3}$CUHK \quad
    $^{4}$HUST \quad
    \\
}

\begin{document}

\maketitle

\begin{abstract} \label{sec:abs}
This work explores enabling Chain-of-Thought~(CoT) reasoning to link visual cues across multiple images.
A straightforward solution is to adapt rule-based reinforcement learning for Vision-Language Models~(VLMs).
However, such methods typically rely on manually curated question-answer pairs, which can be particularly challenging when dealing with fine-grained visual details and complex logic across images.
Inspired by self-supervised visual representation learning, we observe that images contain inherent constraints that can serve as supervision.
Based on this insight, we construct image triplets comprising two augmented views of the same image and a third, similar but distinct image.
During training, the model is prompted to generate a reasoning process to compare these images~(\textit{i.e.}, determine same or different). Then we optimize the model with rule-based reinforcement learning.
Due to the high visual similarity and the presence of augmentations, the model must attend to subtle visual changes and perform logical reasoning to succeed.
Experiments show that, although trained solely on visual comparison tasks, the learned reasoning ability generalizes effectively to a wide range of questions.
Without relying on any human-annotated question-answer pairs, our method achieves significant improvements on multi-image reasoning benchmarks and shows strong performance on general vision tasks.  
\end{abstract}

\section{Introduction}\label{sec:intro}

Making visual analysis with multiple images is crucial in many real-world applications. For example, we understand actions through sequential images or videos, gain 3D awareness by recognizing multiview images, and analyze events by observing differences between states, \textit{etc}.
Although Vision Language Models~(VLMs)~\cite{qwen2.5vl,chen2024internvl,llava,gpt4o,gpt4} demonstrate promising capabilities in understanding single images, we find them struggle to link visual cues across multiple images.

Multi-image understanding requires not only identifying fine-grained visual cues but also performing logical reasoning to uncover correspondences and differences among images.
Recently, reasoning in language models~\cite{deepseekr1,o1,team2025kimi1.5,seedthinker} has been significantly improved through the use of Chain-of-Thought~(CoT) prompting, especially when combined with rule-based reinforcement learning~\cite{grpo}.
Therefore, a straightforward idea to improve multi-image understanding is to extend this reinforcement learning paradigm to the visual domain. However, GRPO~\cite{grpo} requires constructing question-answer pairs with standard answers to compute rewards, which is particularly challenging for tasks involving fine-grained visual details and complex logic across images.

Instead of focusing on constructing QA pairs, we explore how to incentivize VLMs to perform multi-image reasoning with minimal data preparation cost.
Modern VLMs already possess strong perceptual and multimodal capabilities. Meanwhile, recent advances in RL-based single image reasoning~\cite{ThinkLite, MMEureka, r1v} suggest that reasoning ability can be effectively acquired with limited data. 
However, most of these methods still rely on task-specific supervision, such as hand-crafted QA pairs.
To reduce the reliance on manual annotations, we draw inspiration from self-supervised visual representation learning~\cite{SimCLR,Moco,mae,dino}, where images are used as their own source of supervision.
Contrastive learning methods~\cite{Moco,SimCLR}, for instance, learn discriminative representations by pulling together features from different views of the same image and pushing them away from those of different images.
Guided by this principle, we exploit inherent constraints in images as a supervision signal for reward calculation, and present a novel method, \method~(\textbf{M}ultiple \textbf{i}mage \textbf{Con}trast).

Specifically, we construct training triplets consisting of two augmentations of the same image and a third, different but similar image with its own augmentation.
We prompt the VLM to output the thinking process and make comparisons among these images to answer same/different. 
Multiple trajectories are sampled per example, and reinforcement learning is applied using advantages computed from the correctness of the final answer.
A key aspect of our approach is the design of challenging image comparisons. If negative samples are too distinct, the reasoning is trivial.
We address this by sampling frames from the same video or using image editing datasets to find similar images, ensuring subtle differences that require careful visual inspection and reasoning.
Beyond this contrastive framework, we also introduce \textit{Augmented GRPO}, a training strategy that samples trajectories using weak augmentations and optimizes them under stronger augmentations.
This design allows high-quality CoTs to generalize to more difficult images.

Although the model is trained solely on the image comparison task, the learned ability to link visual cues across multiple images generalizes to a wider scope of scenarios. 
For example, the model can predict plausible future actions by analyzing visual changes across frames, distinguish object identities by comparing fine-grained appearance details, or detect subtle camera movement in scene transformations.
Moreover, the contrastive learning process encourages attention to fine-grained details, which also benefits certain single-image understanding tasks like fine-grained layout/attribute understanding. 
Experimental results show that \method achieves strong performance for multi-image understanding~\cite{vlm2bench,muirbench,blink}, and also brings improvements on general vision tasks~\cite{,hallusionbench,MMStar,yue2024mmmu}. 
\vspace{-7pt}

\section{Related Work}\label{sec:related}
\vspace{-5pt}
\noindent \textbf{Vision language model reasoning.} 
Recent studies show that reasoning-capable LLMs~\cite{o1, deepseekr1, team2025kimi1.5, seedthinker} can be effectively guided to generate long CoT~\cite{cot} reasoning processes through reinforcement learning, leading to significant progress on tasks involving complex logic.
Building on these advances, a surge of recent works~\cite{r1v, tan2025reasonrft, peng2025lmmr1, peng2025skyworkr1} has extended CoT reasoning into the vision-language domain.
For example, MM-Eureka~\cite{MMEureka} expands training data coverage across domains and refines RL training strategies.
NoisyRollout~\cite{noisyrollout} introduces image augmentations to enrich the exploration space for policy optimization.
LVAA-Thinking~\cite{VLAAThinker} provides a detailed analysis of supervised fine-tuning and RL for visual reasoning, along with a curated dataset for related tasks.
ThinkLite~\cite{ThinkLite} further improves data efficiency via sample selection with Monte Carlo Tree Search.
While these methods rely heavily on curated training data generated by existing models or human annotations, our work explores an alternative: leveraging inherent constraints within visual data to naturally elicit reasoning ability—without explicit question-answer supervision.

\noindent \textbf{Multi-image understanding.} 
Understanding multiple images is crucial in real-world scenarios that require comparing object states, tracking actions, or recognizing objects across views.
Recent large VLMs~\cite{qwen2.5vl, llavainterleave, llavaonevision, gpt4, gpt4o} have begun to support multi-image inputs natively.
LLaVA-Interleave~\cite{llava} extends LLaVA~\cite{llava} to process interleaved multimodal inputs.
VISC~\cite{zhang2025weaving} introduces focus-centric data to enhance visual reasoning.
Meanwhile, numerous benchmarks~\cite{muirbench, blink, meng2024mmiu, liu2024mmdu, MIRB, MICBench} have been proposed to evaluate multi-image understanding from various angles.
Despite these developments, recent evaluations~\cite{vlm2bench, CoLVA} highlight persistent limitations: VLMs often fail to link fine-grained visual cues across images, such as identifying the same object under different views or detecting subtle state changes for predictive reasoning.
Our work addresses this gap by incentivizing the model to compare the fine details across images and make logical analysis.  
%\vspace{-10pt}

\begin{figure}[t]
\centering 
\includegraphics[width=0.99\linewidth]{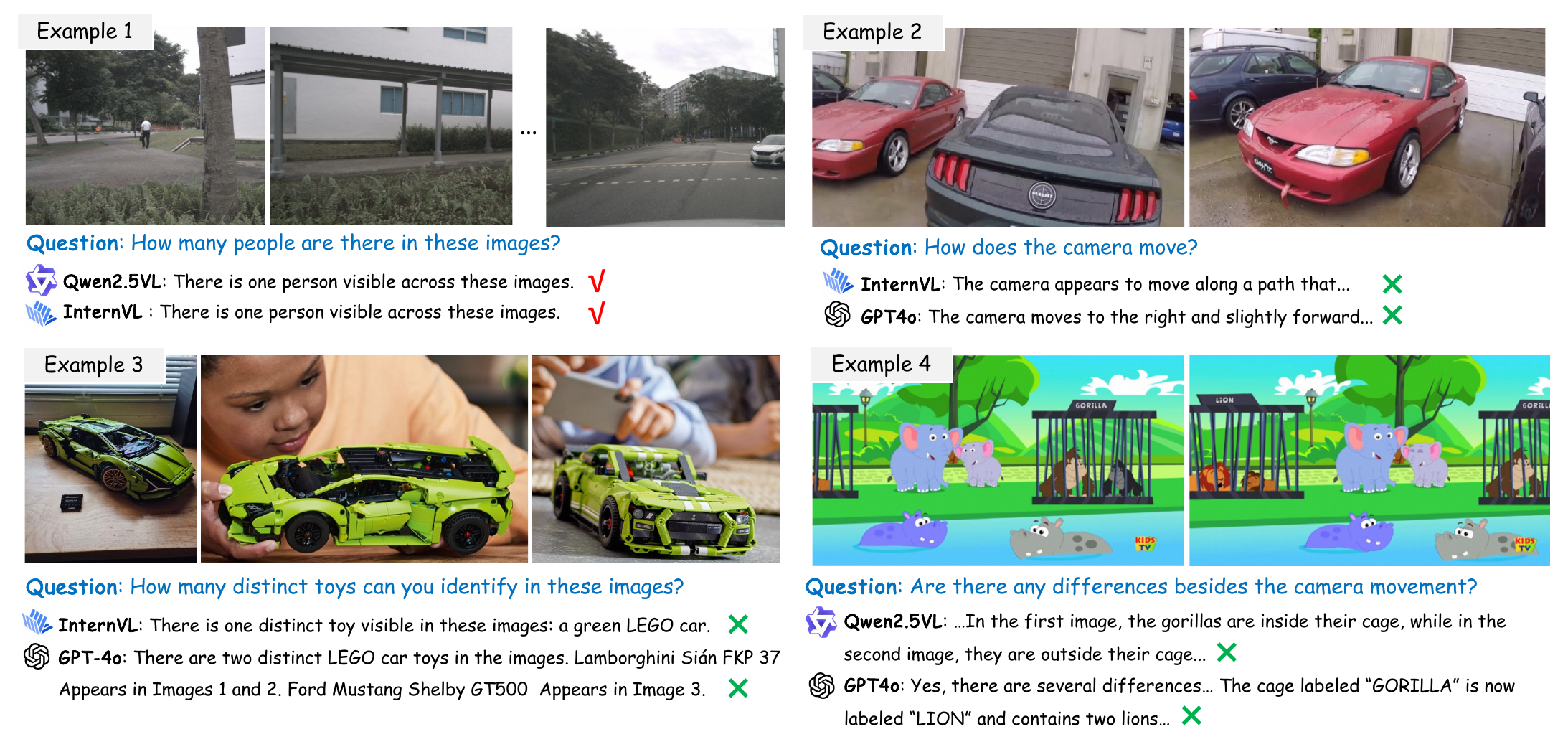} 
\vspace{-5pt}
\caption{%
    \textbf{Challenges for multi-image understanding.} 
    While recent works support multiple images as input, most of them focus on scenarios where each image can be interpreted independently~(\textit{e.g.}, Example~1), which remains relatively easy for current state-of-the-art VLMs. 
    However, many real-world tasks~(\textit{e.g.}, Example~2-4) require models to compare subtle visual differences, align visual cues across images, and reason about object correspondences—capabilities that current VLMs still struggle with. 
    We gather Example~1 from MuirBench~\cite{muirbench}, Example~3 from VLM2-bench~\cite{vlm2bench}, Example~2,4 from real world samples.
}
\label{fig:pilot}
\end{figure}

\section{Method}\label{sec:method}
\subsection{Pilot Study for Multi-image Understanding}
We begin with a pilot study to assess how well current VLMs understand multiple images. As shown in \cref{fig:pilot}, we present examples that highlight the capabilities of several state-of-the-art VLMs, Qwen2.5-VL~\cite{qwen2vl}, InternVL~\cite{chen2024internvl}, and GPT-4o~\cite{gpt4o}. 
While many recent models and benchmarks~\cite{meng2024mmiu,muirbench,MIRB} support multi-image or video inputs, they primarily focus on scenarios like Example~1, where each image can be understood in isolation.
In Example~1, models correctly identify a single person across images, reflecting solid basic perception. 
However, when we examine more complex cases in \cref{fig:pilot}, we observe that VLMs often suffer from severe hallucinations.
In Example~2, both models fail to infer the correct camera movement, showing weaknesses in spatial reasoning. In Example~3, VLMs cannot distinguish between different car toys, indicating difficulty with cross-image comparison. 
Example~4 further reveals failure in tracking semantic changes across images, with hallucinated object positions and misidentified labels. These cases highlight that current VLMs still lack robust visual comparison abilities essential for multi-image understanding.
These examples typically require the model to explicitly link visual cues across images, analyze fine-grained differences, and reason about inter-image correspondences.

As current VLMs already possess strong abilities in single-image perception~(\textit{e.g.}, reading fine-grained text) and demonstrate solid commonsense knowledge, as evidenced by their performance on standard vision benchmarks. We hypothesize that their primary limitation in multi-image understanding lies in their inability to compare and connect visual information across images.
To address this gap, we focus on enhancing the meta-cognitive ability of \textbf{visual comparison}, the core skill needed for effective multi-image reasoning.

\subsection{Multi-image Contrast}

Rather than collecting data for each specific multi-image task, we aim to improve VLMs' general capacity to analyze and reason over multiple images by targeting the core meta skill: \textbf{visual comparison}.
Inspired by the principles of self-supervised learning, we design a lightweight and scalable framework that encourages the model to distinguish similar yet distinct images.
By simulating contrastive visual situations and prompting the model to generate structured reasoning trajectories, we aim to enhance its ability to perceive fine-grained differences, establish correspondences, and perform step-by-step comparisons across images.

Here, we elaborate on the pipeline of \method. The overall framework consists of the following main steps. First, we identify and construct contrastive image samples that are visually similar yet different. Then, we apply data augmentation to build informative training triplets. Finally, we leverage Augmented GRPO to evaluate a set of reasoning trajectories and optimize the VLM accordingly.

\noindent \textbf{Image selection.}  
We begin by selecting image pairs that are visually similar but exhibit subtle differences, which serve as contrastive supervision signals. We denote such a pair as $(I_a, I_b)$, where $I_a$ and $I_b$ are distinct images sharing high structural similarity~(\textit{e.g.}, similar layout or background), but with small detail variations.

We leverage two types of data sources that naturally fulfill this requirement: video frames and image editing datasets.  
For video data, we randomly sample $(I_a, I_b)$ from the same video with a temporal gap of 2 seconds, and compute their Structural Similarity~(SSIM) to filter out near-identical pairs.
For image editing data, each $(I_a, I_b)$ pair consists of a "before" and "after" edited image. We compute the pixel-wise Mean Squared Error~(MSE) to remove significantly different pairs. These constraints ensure that the collected pairs exhibit subtle but meaningful changes.

\noindent \textbf{Image augmentation.} 
While the selected image pairs already exhibit subtle variations, directly learning to distinguish them may still lead to shortcut learning. To increase task complexity and encourage detailed reasoning, we apply data augmentation to create diverse image views.

\begin{figure}[t]
\centering 
\includegraphics[width=0.99\linewidth]{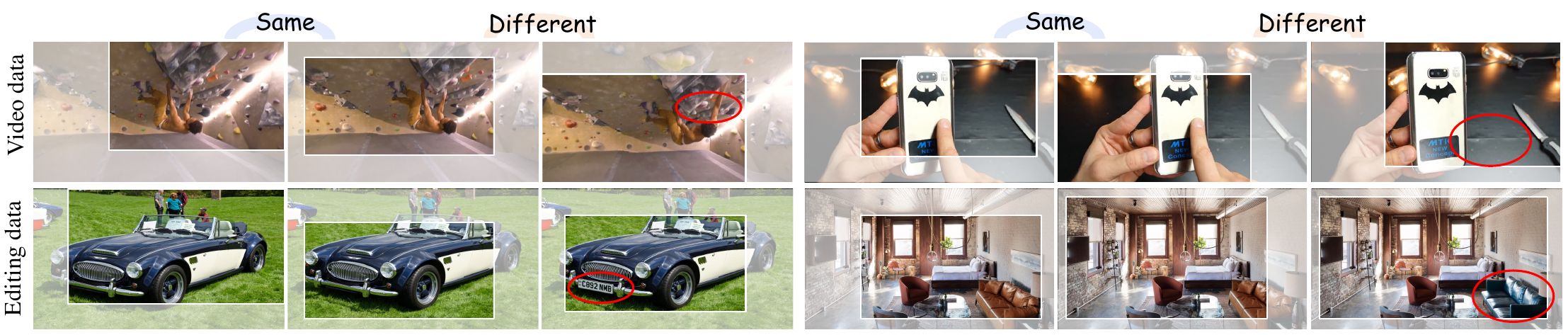} 
\vspace{-5pt}
\caption{%
    \textbf{Demonstrations for contrastive samples.} The first row shows two triplets from the video, and the second row demonstrates samples from image editing datasets. 
    These samples are visually similar but contain subtle differences~(marked with red circles), on which we apply random cropping and resizing.  
    In each triplet, the first two images are the same, and the third image is different. 
}
\label{fig:data}
\end{figure}

As the visualization examples provided in \cref{fig:data}, given a source image $I$, we generate two augmented versions via random cropping and resizing~(they do not change the content of images). 
For each image pair $(I_a, I_b)$, we thus construct a contrastive triplet:
\[
\mathcal{T} = \{ \mathcal{T}_1(I_a), \mathcal{T}_2(I_a),  \mathcal{T}_3(I_b) \} ,
\]

\subsection{Augmented GRPO}\label{sec:augmented}
\noindent \textbf{Question-answer formulation.} 
After getting the triplets that contain similar images and their augmentations. We construct QA pairs for reinforcement learning. Given an image triplet, we add reasnoning prompt and user questions as follows:

\begin{tcolorbox}[
    colback=gray!10!white,     % 浅灰色背景
    colframe=black,            % 黑色边框
    boxrule=1pt,               % 边框线粗细
    boxsep=4pt,                % 内容与边框的内边距（上下左右）
    top=4pt, bottom=4pt,       % 上下边距略增
    left=8pt, right=8pt        % 左右间距扩大
]
\begin{center}
\textbf{Reasoning Template of MiCo}
\end{center}
\textbf{Reasoning Prompt:} First output the thinking process in <think> </think> and give the final answer in <answer> </answer> tags. \\

\textbf{User Question:} Regardless of the augmentation, are image1 and image2 the same? How about image2 and image3, image1 and image3? Only return T(True) or F(False) in <answer> </answer>, for example <think> </think> <answer>TFT</answer>.\\
\end{tcolorbox}

To increase the diversity and balance the difficulties of questions, besides the image triplet, we also construct image pairs and design the corresponding prompts for comparing two images. In addition, we use GPT-4o~\cite{gpt4o} to expand the user question of the same meaning but with various expressions.

\noindent \textbf{Rollout augmentation.} 
For each question $q$ with augmented images $\mathcal{T}$, the original GRPO samples a group of outputs $\{o_1, o_2, \cdots, o_G\}$ from the old policy $\pi_{\theta_{old}}$ and then optimizes the policy model $\pi_{\theta}$.
To better leverage difficult samples that arise from strong augmentations, we sample trajectories using weakly augmented inputs $\mathcal{T}^{w}$, which are easier to produce valid reasoning chains. These sampled trajectories are then used to optimize the policy on stronger augmented contexts $\mathcal{T}^{s}$, effectively transferring reliable behavior to harder instances.

\noindent \textbf{Training objective.} The training objective of Augmented GRPO could be formulated as follows. 
This objective encourages the policy to assign higher likelihoods to responses with higher relative rewards within each group. 

\begin{equation}
\begin{split}
    \mathcal{J}(\theta) &= \mathbb{E}{[q \sim P(Q},\ \{o_i\}_{i=1}^G \sim \pi_{\theta_{old}}(O|q)] \\
    = & \frac{1}{G}\sum_{i=1}^G \left( \min \left( \frac{\pi_\theta(o_i |q)  }{\pi_{\theta_{old}}(o_i |q) } A_i,\ \text{clip} \left( \frac{\pi_\theta(o_i |q)  }{\pi_{\theta_{old}}(o_i |q)  }\right) A_i \right) - \beta\ \mathbb{D}_{KL}\left(\pi_{\theta} || \pi_{ref}\right)\right) ,
\end{split}
\label{eq:GRPO-obj}
\end{equation}

\begin{equation}
    \mathbb{D}_{KL}\left(\pi_{\theta} || \pi_{ref}\right) = \frac{\pi_{ref}(o_i \mid q)  }{\pi_{\theta}(o_i \mid q) }
    - \log\frac{\pi_{ref}(o_i \mid q) }{\pi_{\theta}(o_i \mid q) }
    - 1,
\end{equation}

where $\epsilon$ and $\beta$ are hyperparameters. Following GRPO~\cite{grpo},  $A_i$ is the normalized advantage computed based on rewards $\{r_1, r_2, \ldots, r_G\}$.

\begin{equation}
    A_i = \frac{r_i - {\mathrm mean(\{r_1, r_2, \cdots, r_G\})}}{{\mathrm std(\{r_1, r_2, \cdots, r_G\})}}.
\end{equation}

Following DeepSeek-R1~\cite{deepseekr1}, we leverage the binary format reward and the accuracy reward, which considers the matching of ``<think> </think> <answer> </answer>'' tags, and the correctness of the final answer. For the triplet comparisons, we get 1 for the accuracy reward only if we make correct comparisons for all three pairs.

\begin{algorithm}[t]
\caption{\method: Reinforcement Multi-image Reasoning}
\label{alg:agrpo}
\begin{algorithmic}[1]
\State \textbf{Input:} Policy $\pi_\theta$, old policy $\pi_{\theta_{\text{old}}}$, image triplet dataset $\mathcal{D} = \{(I_1, I_2, I_3)\}$, training steps $T_{\text{max}}$, group size $G$, clip parameter $\epsilon$, weak augment operators $\mathcal{T}^{\text{w}}$, strong augment operators $\mathcal{T}^{\text{s}}$
\For{$t = 1$ to $T_{\text{max}}$}
    \State Sample triplet $(I_1, I_2, I_3) \sim \mathcal{D}$
    
    \State Apply weak augmentation: $(I_1^w, I_2^w, I_3^w) = \mathcal{T}^{\text{w}}(I_1, I_2, I_3)$
    \State Apply strong augmentation: $(I_1^s, I_2^s, I_3^s) = \mathcal{T}^{\text{s}}(I_1, I_2, I_3)$
    
    \State Construct prompts $\mathbf{q}^w$ and $\mathbf{q}^s$ from the weak and strong augmented triplets, respectively
    
    \State Sample $G$ CoT responses $\{\mathbf{o}_i\}_{i=1}^G$ from $\pi_{\theta_{\text{old}}}(\cdot \mid \mathbf{q}^w)$ \Comment{Rollouts from \textcolor{citeblue}{weak prompt}  }
    \State Evaluate reward $R_i = R(I^w, \mathbf{q}^w, \mathbf{o}_i)$ for each $i = 1, \dots, G$
    \State Compute group baseline $\bar{R} = \frac{1}{G} \sum_{i=1}^{G} R_i$, and advantages $\hat{A}_i = \frac{R_i - \bar{R}}{\sigma(R)}$
    
    \State Optimize $\pi_\theta$ on the \textcolor{citeblue}{strong prompt} $\mathbf{q}^s$ using the group rollouts:
    \State $L(\theta) = \frac{1}{G} \sum_{i=1}^{G} \min\left(r_i \hat{A}_i, \mathrm{clip}(r_i, 1-\epsilon, 1+\epsilon)\hat{A}_i\right)$, where $r_i = \frac{\pi_\theta(\mathbf{o}_i \mid \mathbf{q}^s)}{\pi_{\theta_{\text{old}}}(\mathbf{o}_i \mid \mathbf{q}^s)}$
    
    \State $\theta \leftarrow \theta - \nabla_\theta L(\theta)$
    \State $\theta_{\text{old}} \leftarrow \theta$
\EndFor
\end{algorithmic}
\end{algorithm}

\noindent \textbf{Overall algorithm.} Our \method could be summarized in \cref{alg:agrpo}. We first construct an image triplet consisting of two augmented views of the same image and a third, visually similar but distinct image~(with augmentations). The model is prompted to perform multi-image comparison and generate reasoning trajectories. During training, 
chain-of-thought responses are sampled from the weakly augmented views, and the policy is optimized on the strongly augmented ones using rule-based reinforcement learning. This process enables the model to learn fine-grained visual reasoning in a self-supervised manner.

\section{Experiments}\label{sec:exp}

\subsection{Implementation Details}
\noindent \textbf{Hyper-parameters.} For the baseline model, we follow previous works~\cite{MMEureka,noisyrollout,ThinkLite,VLAAThinker} and select Qwen2.5-VL-7B~\cite{qwen2.5vl}. 
For the training data, we use OmniEdit~\cite{wei2024omniedit} for image editing pairs and extract video frames from Vidgen-1M~\cite{tan2024vidgen}. 
The part of reinforcement learning follows GRPO~\cite{grpo}, we set a format reward and accuracy reward with the weight of 1:1, respectively. Besides, we also apply a KL regularization with a weight of 0.01. 
During training, we follow previous works~\cite{MMEureka} to skip the rollout group with all correct/false answers.
During training, we use a learning rate of 1e-6 and set the batch size of 16. For each training sample, we generate a group of 8 rollouts. We train the model for 600 iterations on 8$\times$A100 GPUs.

\noindent \textbf{Evaluation protocols.} For evaluation, we follow the default hyper-parameters of Qwen2.5-VL~\cite{qwen2vl} and utilize the VLMEvalKit~\cite{duan2024vlmevalkit}. For reasoning baselines, we adopt their official prompting formats. Minor inconsistencies in results may occur due to differences in the implementation details of evaluation frameworks or answer parsing logic.

\begin{table}[t]
\caption{\textbf{Performance on VLM2-Bench}~\cite{vlm2bench}, which evaluates the ability to compare and link fine-grained visual cues across multiple images.  
Without relying on any human- or model-annotated data, \method achieves significant improvements and sets a new state-of-the-art.  
Reasoning-based models (marked with $\bullet$) are evaluated using their corresponding prompting strategies.}
\vspace{-2pt}
\centering
\footnotesize
\scalebox{0.77}{
\renewcommand{\arraystretch}{1.3}
\begin{tabular}{l|cc|ccc|cccc|cc}
\hline
{Baselines or Models} & \multicolumn{2}{c|}{{General}} & \multicolumn{3}{c|}{{Object}} & \multicolumn{4}{c|}{{Person}} & \multicolumn{2}{c}{{Overall*}} \\
 & Mat & Trk & Cpr & Cnt & Grp & Cpr & Cnt & Grp & VID & Avg & $\Delta_{\text{human}}$ \\
\hline
Chance-Level         & 25.00 & 25.00 & 50.00 & 34.88 & 25.00 & 50.00 & 34.87 & 25.00 & -    & 32.73 & -61.44 \\
Human-Level          & 95.06 & 98.11 & 96.02 & 94.23 & 91.29 & 97.08 & 92.87 & 91.17 & 100.00 & 95.16 & 0.00 \\
\hline
$\circ$ LLaVA-OneVision\cite{llavaonevision}   & 16.60 & 13.70 & 47.22 & 56.17 & 27.50 & 62.00 & 46.67 & 37.00 & 47.25 & 39.35 & -55.81 \\

$\circ$ LLaVA-Video-7B~\cite{llavavideo}       & 18.53 & 12.79 & 54.72 & 62.47 & 28.50 & 62.00 & 66.91 & 25.00 & 59.00 & 45.65 & -49.51 \\

$\circ$ LongVQA-7B~\cite{longva}           & 14.29 & 12.98 & 46.53 & 49.47 & 29.00 & 58.00 & 41.56 & 25.00 & 45.00 & 37.10 & -58.06 \\

$\circ$ mPLUG-Owl2-7B~\cite{mplugowl2}        & 17.37 & 18.26 & 49.17 & 62.97 & 31.00 & 63.00 & 58.06 & 29.00 & 43.00 & 40.87 & -54.31 \\

$\circ$ Qwen2-VL-7B~\cite{qwen2.5vl}          & 18.07 & 19.18 & 68.08 & 61.84 & 37.50 & 72.00 & 67.92 & 47.00 & 55.25 & 49.76 & -45.40 \\

$\circ$ InternVL2.5-8B~\cite{chen2024internvl}       & 41.24 & 26.53 & 72.22 & 67.65 & 40.00 & \textbf{85.00} & 66.67 & 52.25 & 50.25 & 55.41 & -39.75 \\

$\circ$ InternVL2.5-26B~\cite{chen2024internvl}      & 30.50 & 30.59 & 43.33 & 51.48 & 52.50 & 59.50 & 59.67 & 61.25 & 45.25 & 45.59 & -49.57 \\

$\circ$ Qwen2.5-VL-7B~\cite{qwen2.5vl}        & 35.91 & 43.38 & 71.39 & 41.72 & 47.50 & 80.00 & 59.76 & 69.00 & 45.00 & 54.82 & -40.34 \\

$\circ$ GPT-4o~\cite{gpt4o}               & 37.45 & 39.27 & 74.17 & \textbf{80.62} & 57.50 & 50.00 & \textbf{90.50} & 47.00 & \textbf{66.75} & 60.36 & -34.80 \\
\hline

$\bullet$ MM-Eureka-7B~\cite{MMEureka}     & 55.60  & 47.03	& 74.10	&52.50	& 54.00	 & 77.50 & 60.00 & 51.00 & 43.50 & 57.24 &-37.91 \\

$\bullet$ NoisyRollout-7B~\cite{noisyrollout}            &   40.93 & 43.83 & 63.33 & 50.83 & 34.50 & 70.50 & 63.33 & 47.00 & 36.50 & 50.08 & -45.08 \\

$\bullet$ ThinkLite-VL-7B~\cite{ThinkLite}           &    40.45 & 46.58 & 75.56 & 62.50 & 49.50 & 77.50 & 62.50 & 51.00 & 36.50 & 55.79 & -39.37 \\

$\bullet$ VLAA-Thinker-7B~\cite{VLAAThinker}               & 47.49 & 63.03 & 72.20 & 61.40 & 55.00 & 71.00 & 57.50 & 51.00 & 47.75 & 58.49 & -36.67 \\

\hline

$\circ$ Qwen2.5-VL-7B-CoT\cite{qwen2.5vl}   & 43.24	 & 42.92  & 66.39	& 50.56	& 36.00 & 62.50	 & 55.83 & 39.00 & 36.75 &	48.91 &	-46.24 \\

$\bullet$ \method-7B-CoT  & \textbf{57.14} & \textbf{67.12} & \textbf{81.94} & 56.67 & \textbf{58.00} & 65.00 & 57.50 & 62.00 & 44.25 & \textbf{61.06} & \textbf{-34.09} \\

\rowcolor[gray]{0.9} \small
$\Delta$ Improvement  & +13.90 & +24.20 & +15.55 & +6.11 & +22.00 & +2.50 & +1.67 & +23.00 & +7.50 & +12.93 & +12.93 \\

\hline
\end{tabular}
}
\label{tab:vlm2bench}
\vspace{-5pt}
\end{table}

\subsection{Result Analysis for Multi-image Comparison}
\noindent \textbf{Evaluation metrics.} 
We first report the model performance on VLM2-Bench~\cite{vlm2bench}. This benchmark mostly aligns with our intention of linking fine-grained visual cues across images. 
Specifically, VLM2-Bench includes three tracks: General Cue~(GC), Object-centric Cue~(OC), and Person-centric Cue~(PC). Each track consists of subtasks with specific metrics: Mat~(Matching) and Trk~(Tracking) use paired T/F accuracy; Cpr~(Comparison) evaluates consistency by requiring the model to correctly answer both a positive and its corresponding negative statement; Cnt~(Counting) uses normalized error to measure numerical prediction accuracy; Grp~(Grouping) is a multiple-choice task assessing clustering ability; and VID~(Video Identity Describing) is scored based on GPT-4o evaluation of open-ended descriptions.

\noindent \textbf{Result analysis.}
As shown in \cref{tab:vlm2bench}, we present the comparison results on VLM2-Bench. We observe that all existing open- and closed-source models lag behind human performance by a large margin. Among them, GPT-4o~\cite{gpt4o} demonstrated clear advantages over other models.
Thanks to the strong generalization ability of reinforcement learning, recent reasoning VLMs~\cite{MMEureka,noisyrollout,ThinkLite,VLAAThinker} have shown consistent improvements when built upon Qwen2.5-VL-7B~\cite{qwen2.5vl}.
We report the performance of \method in the final block. Trained with contrastive triplets, \method effectively learns the core ability to compare images, achieving substantial gains across multiple tasks and obtaining the best average performance overall. Notably, our 7B model even outperforms GPT-4o.
However, we find that CoT reasoning does not benefit all sub-tasks equally. Specifically, for tasks involving human faces (person track), CoT-based models offer limited or even negative gains compared to no-CoT counterparts. We hypothesize that human identity representations, such as facial nuances, are difficult to verbalize, thus limiting the benefit of language-based reasoning.
In contrast, object-level identity differences~(\textit{e.g.}, logos, textures, shapes) are more readily describable, allowing CoT reasoning to help reduce hallucinations and improve distinction.

\begin{table}[t]
\fontsize{7.5pt}{10pt}\selectfont
\centering
\caption{\textbf{Ablation studies} on key configurations. We conduct experiments on VLM2-Bench~\cite{vlm2bench} and report the average accuracy across the general, object, and person tracks.  
For each ablation, all other settings are kept consistent with our final model to ensure fair comparisons.}
\vspace{-5pt}
\begin{minipage}[t]{0.48\linewidth}
    \centering
    {(a)~Learning Paradigm}
    \scalebox{0.99}{
    \renewcommand{\arraystretch}{1.2}
    \begin{tabular}{lccc}
        \hline
        & General & Object & Person \\
        \hline
        Qwen2.5-VL~\cite{qwen2.5vl}  & 39.64   &	53.53  &	63.44  \\
        SFT      & 42.90	& 51.15	 & 55.98    \\
        No-CoT RL  & 45.36	& 50.01	 & 55.23    \\
        CoT RL    & 62.13	 & 65.53 	  &  57.18   \\
        \hline
    \end{tabular}
    }
\end{minipage}
\hspace{5pt}
\begin{minipage}[t]{0.48\linewidth}
    \centering
    {(b)~Data Source}
    \scalebox{0.99}{
    \renewcommand{\arraystretch}{1.2}
    \begin{tabular}{lcccc}
        \hline
        & General & Object & Person  \\
        \hline
        Edit Data$^1$~\cite{wei2024omniedit}  & 61.23	& 65.33	& 56.35 \\
        Edit Data$^2$~\cite{zhao2024ultraedit}  & 60.88 & 64.33 & 55.27 \\
        Video Data  &  60.29	& 64.50	& 55.68  \\
        Edit$^1$ + Video   & 62.13	 & 65.53 	  &  57.18   \\
        \hline
    \end{tabular}
    }
\end{minipage}

\vspace{7pt}
\begin{minipage}[t]{0.48\linewidth}
    \centering
    {(c)~Rollout Augmentation}
    \scalebox{0.99}{
    \renewcommand{\arraystretch}{1.2}
    \begin{tabular}{lccc}
        \hline
        & General & Object & Person  \\
        \hline
        Qwen2.5-VL~\cite{qwen2.5vl}   & 39.64   &	53.53  &	63.44  \\
        (Strong, Strong)   & 59.41  & 64.00	 & 56.98 \\
        (Weak, Weak)    & 55.58	  & 62.03  & 54.81 \\
        (Weak, Strong)    & 62.13	 & 65.53 	  &  57.18   \\
        \hline
    \end{tabular}
    }
\end{minipage}
\hspace{5pt}
\begin{minipage}[t]{0.48\linewidth}
    \centering
    {(d)~Sample Formulation}
    \scalebox{0.99}{
    \renewcommand{\arraystretch}{1.2}
    \begin{tabular}{lccc}
        \hline
        & General & Object & Person  \\
        \hline
        Qwen2.5-VL~\cite{qwen2.5vl}   & 39.64   &	53.53  &	63.44  \\
        Image Pairs  & 56.41 &	66.98 &	55.68  \\
        Image Triplets  & 60.64 &	65.33	& 55.81 \\
        Pairs + Triplets    & 62.13	 & 65.53 	  &  57.18   \\
        \hline
    \end{tabular}
    }
\end{minipage}

\vspace{5pt}
\begin{minipage}[t]{0.48\linewidth}
    \centering
    {(e)~Prompt Diversity}
    \scalebox{0.99}{
    \renewcommand{\arraystretch}{1.2}
    \begin{tabular}{lccc}
        \hline
        & General & Object & Person \\
        \hline 
        Qwen2.5-VL~\cite{qwen2.5vl}   & 39.64   &	53.53  &	63.44  \\
        Single Prompt & 55.53	 & 64.16	& 51.50 \\
        20 Variations   & 62.13	 & 65.53 	  &  57.18   \\
        50 Variations   & 63.13	 & 65.29  & 54.93    \\
        \hline
    \end{tabular}
    }
\end{minipage}
\hspace{5pt}
\begin{minipage}[t]{0.48\linewidth}
    \centering
    {(f)~Image Augmentations}
    \scalebox{0.99}{
    \renewcommand{\arraystretch}{1.2}
    \begin{tabular}{lccc}
        \hline
        & General & Object & Person  \\
        \hline
        Base~\tiny{(Crop, Resize)}  & 62.13 & 65.53  &  57.18   \\
        Base + Flip  & 61.13 &	63.98 &	56.77  \\
        Base + Rotat.  & 62.58  & 65.03 &	55.86    \\
        Base + Color.  & 60.15	& 64.26	 & 54.86  \\
        \hline
    \end{tabular}
    }
\end{minipage}
\label{tab:ablation}
\end{table}

\subsection{Ablation Studies}

We conduct a series of ablation studies to validate the effectiveness of our core designs. As an initial exploration of visual reasoning, we also analyze the impact of some basic configurations.

\noindent \textbf{Training strategies. }
In \cref{tab:ablation}~(a), we evaluate different training paradigms.
We first apply supervised fine-tuning~(SFT) on our contrastive dataset, allowing the model to directly predict the final answer. We observe that this leads to minor gains on the general track, which more closely aligns with the training task. However, the ability acquired through SFT does not generalize well to more diverse reasoning tasks.    
We also test a no-CoT reinforcement learning baseline, where the model is trained to output answers directly using GRPO~\cite{grpo}. Due to the absence of intermediate reasoning steps, the resulting trajectories are short and behave similarly to SFT, yielding limited improvements.

\noindent \textbf{Data source. }
In \cref{tab:ablation}~(b), we compare different training data sources.
Both the image editing data (OmniEdit~\cite{wei2024omniedit}) and video-derived frames (VidGen~\cite{tan2024vidgen}) individually support effective learning. Combining these two heterogeneous sources further enhances performance. We also validate that our framework is not tied to specific editing styles, as models trained on either OmniEdit or UltraEdit~\cite{zhao2024ultraedit} generalize well, demonstrating robustness to the editing domain variation.

\noindent \textbf{Rollout augmentation. } In this work, we leverage weak augmentations for rollout sampling, and use these high-quality answers to optimize harder questions with stronger augmentations. In \cref{tab:ablation}~(c), we report different combinations of augmentation in ``(sampling, optimization)'' process. We show that, strong augmentations are vital for contrastive learning compared with weak augmentations, and our rollout augmentation strategy gets the best performance. 

\noindent \textbf{Sample formulation. } As discussed in \cref{sec:augmented}, we construct prompts based on either image pairs or image triplets.
While we initially suspected that binary image-pair comparisons (with 50\% guess probability) might result in low-quality CoTs, our experiments reveal that they still contribute positively to performance.
In practice, we find that combining both formats—pair-based and triplet-based leads to the best results.

\noindent \textbf{Other configurations. } In \cref{tab:ablation}~(e),~(f), we explore the effects of prompt and augmentation diversity.
We observe that increasing the variation of image prompts helps prevent overfitting, with performance saturating at around 50 distinct prompt templates. For image augmentations, we experimented with various techniques and ultimately selected random cropping and resizing as the default setting based on empirical performance.

\begin{table}[t]
\caption{%
    \textbf{Quantitative results on general vision benchmarks.} We report performance for wide scenarios. Multi-image benchmarks are marked in bold. \method brings steady improvements compared with our baseline, and gets competitive results against other visual reasoning models.
}
\label{tab:general_bench}
\vspace{-3pt}
\centering\footnotesize
\setlength{\tabcolsep}{4pt}
\scalebox{0.85}{
\renewcommand{\arraystretch}{1.2}
\begin{tabular}{lcccccc}
\hline
& \textbf{MuirBench}~\cite{muirbench} &  \textbf{BLINK}~\cite{blink} & Hallusion~\cite{hallusionbench}
 & MMStar~\cite{MMStar} &  MMMU~\cite{yue2024mmmu} & MathVistas~\cite{mathvista} \\
\hline

 MM-Eureka-7B~\cite{MMEureka} & 60.57 & 54.39 & 68.45 & 65.73  & 54.11 & 72.00 \\

 NoisyRollout-7B~\cite{noisyrollout}  & 59.61  & 56.07 & 66.66 & 65.66  & 54.55 & 71.60   \\

 VLAA-Thinker-7B~\cite{VLAAThinker}  & 61.00 & 54.81 & 69.08 & 63.60 & 54.44 & 70.80 \\

 ThinkLite-VL-7B~\cite{ThinkLite}  & 57.62 & 55.81 & 72.97  & 66.80 & 53.55 & 71.89 \\

\hline
Qwen2.5VL-7B~\cite{qwen2.5vl} & 58.43 & 55.54 &  69.50 &   64.06 & 54.11 & 67.10 \\

\method-7B  & 60.53 & 57.23 & 69.61 & 65.60  & 54.77 & 67.90 \\

\rowcolor[gray]{0.9} \small
$\Delta$ Improvement & +2.10 & +1.69 & +0.11 & +1.54 & +0.66 & +0.80 \\

\hline
\end{tabular}
}
\end{table}

\begin{table}[t]
\caption{%
    \textbf{Task analysis for visual reasoning.} We list representative sub-tasks from MuirBench~\cite{muirbench} and BLINK~\cite{blink} to analyze the generalization ability and limitations for \method.} 
\label{tab:general_bench}
\vspace{-3pt}
\centering\footnotesize
\setlength{\tabcolsep}{4pt}
\scalebox{0.85}{
\renewcommand{\arraystretch}{1.2}
\begin{tabular}{lcccccc}
\hline
& Visual retrieval & Semantic Corr. & Spatial Rela. & Scene Under. & Forensic Det.  & Relative Depth \\ 
\hline
Qwen2.5VL-7B~\cite{VLAAThinker} & 63.69 & 33.09 &  88.81  & 61.82 &   48.48  & 81.45 \\

MM-Eureka-7B~\cite{MMEureka} & 57.19 \color{red}{+} & 33.09 \color{red}{+}& 82.51 \color{green}{-} & 67.74 \color{red}{++}  & 50.00 \color{red}{+}  &  75.80  \color{green}{-} \\

VLAA-Thinker-7B~\cite{VLAAThinker} & 68.83 \color{red}{+}  & 34.53 \color{red}{+} & 86.71 \color{green}{-} & 69.89 \color{red}{++} & 47.72 \color{green}{-} & 76.61 \color{green}{-} \\

\method-7B  & 71.23 \color{red}{++} & 41.72 \color{red}{++} & 90.20 \color{red}{+} &   63.97 \color{red}{+}  & 47.72 \color{green}{-}  & 78.22 \color{green}{-}  \\
\hline
\end{tabular}
}
\vspace{-10pt}
\end{table}

\subsection{Analysis on General Vision Tasks}
In this section, we evaluate the generalization ability and capacity boundaries of \method on a broader range of vision tasks. We first report quantitative results on additional benchmarks and analyze performance across more diverse task types.

\noindent \textbf{Results on additional benchmarks.}
As shown in \cref{tab:general_bench}, we evaluate \method on MuirBench~\cite{muirbench} and BLINK~\cite{blink}, both of which are representative multi-image understanding benchmarks.
To further assess generalization, we also include several single-image benchmarks, including MMStar~\cite{MMStar}, MMMU~\cite{yue2024mmmu}, HallusionBench~\cite{hallusionbench}, and MathVista~\cite{mathvista}.

Compared to methods trained with manually curated supervision, our contrastive learning framework exhibits strong performance on multi-image understanding tasks, where relational reasoning across images is crucial.
While \method also improves over standard baselines on single-image tasks, its performance remains behind models trained with task-specific guidance, particularly in complex scenarios like visual mathematics, where symbolic reasoning and structured representation are essential but not explicitly modeled in our current training paradigm.

\noindent \textbf{Task-wise analysis.}
We further analyze the performance of \method on specific sub-tasks from MuirBench~\cite{muirbench} and BLINK~\cite{blink} to better understand its strengths and limitations.
Our contrastive learning framework demonstrates clear advantages on correspondence-style tasks, such as \textit{Visual Retrieval} and \textit{Semantic Correspondence}, where \method outperforms other reasoning models. These results highlight the model’s strength in aligning multimodal signals through relational comparisons.
In addition, \textit{Spatial Relation} tasks—which evaluate the model’s understanding of image layout and object positioning—also benefit from contrastive training. By encouraging attention to relative positions among visual entities, \method achieves the highest accuracy in this category.

On the other hand, \method lags behind models trained with manually curated reasoning datasets on tasks such as \textit{Scene Understanding} and \textit{Forensic Detection}, which typically rely on single-image question answering. These tasks often demand domain-specific priors or curated logic patterns that are less emphasized in our data construction process.
We also observe that \textit{Relative Depth}, a representative spatial reasoning task, remains challenging for all models and exhibits a noticeable performance drop. This suggests that depth-aware understanding is not sufficiently captured by current training signals. Incorporating explicit spatial or geometric cues remains a promising direction for future work toward building spatially grounded vision-language models.

\begin{figure}[t]
\centering 
\includegraphics[width=0.99\linewidth]{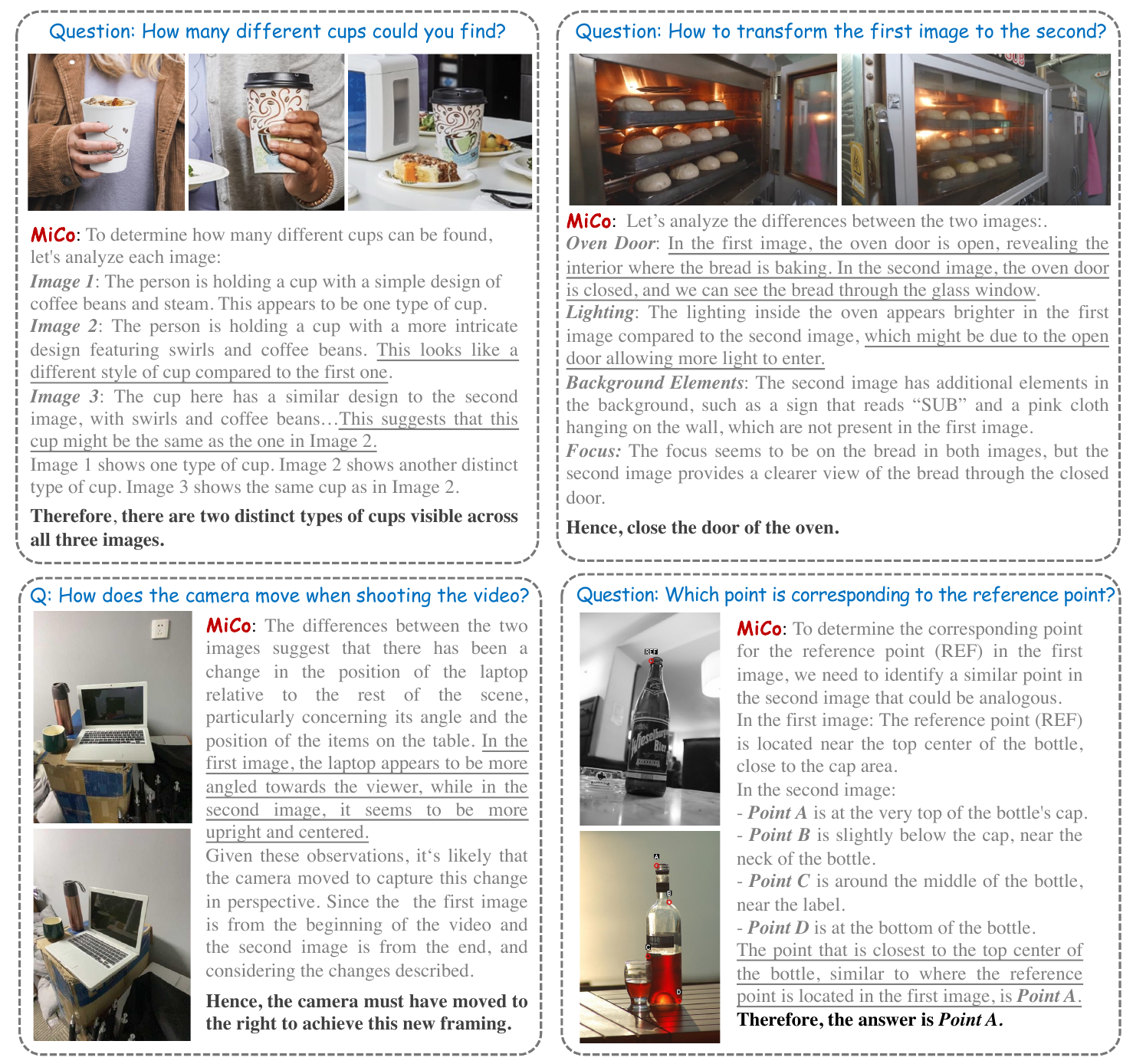} 
\vspace{-8pt}
\caption{%
    \textbf{Demonstrations for visual reasoning.} 
    Given a question, \method first examines the details of each image to identify answer-related visual cues, and then performs cross-image comparisons to derive the final answer.
    The reasoning processes are marked in gray, with key contents underlined.
}
\vspace{-8pt}
\label{fig:demo}
\end{figure}

\vspace{-5pt}
\subsection{Qualitative Analysis}
\vspace{-8pt}
We provide qualitative demonstrations in \cref{fig:demo} to illustrate the multi-image reasoning process of \method. For each question, the model first generates a detailed reasoning chain that carefully compares visual cues across the input images. This includes analyzing fine-grained differences and similarities that are relevant to the question. Based on this reasoning, the model then produces a final answer that successfuly addresses the query objective.

\vspace{-5pt}
\section{Conclusion}\label{sec:conclusion}
\vspace{-10pt}
In this work, we propose a self-supervised framework that leverages inherent image constraints to incentivize multi-image reasoning in VLMs.
We identify that the core challenge lies in linking visual cues across images. To address this, we adopt contrastive learning principles and construct image triplets for reinforcement training.
To further enhance reasoning, we introduce Augmented GRPO, which samples rollouts from simpler examples and optimizes the model on harder ones.
Although trained solely on image comparison tasks, our model generalizes well and achieves strong results across multiple benchmarks.

\noindent \textbf{Limitations and future directions.}
While our approach supports general reasoning through visual comparisons, it shows limited effectiveness on specialized tasks such as face verification, visual math, and spatial understanding, where structured priors or domain-specific knowledge are required. In future work, we plan to explore more efficient data construction strategies tailored to these domains.

\clearpage
{\small
\bibliographystyle{plain}
\bibliography{ref}
}

\appendix

\clearpage

%In the supplementary materials, we first elaborate on more implementation details in \cref{sec:moredetail}. Then, then we report more experimental results in \cref{sec:moreexp}. Afterwards, we give more qualitative examples in \cref{sec:morequa} to demonstrate and analyze the incentivized reasoning process.   Finally, we discuss the potential social impact in \cref{sec:impact}.

\section{More Implementation Details}\label{sec:moredetail}

\noindent \textbf{Prompt expansion.} In the main paper, we give an example of the user question as follows:

\begin{tcolorbox}[
    colback=gray!10!white,     % 浅灰色背景
    colframe=black,            % 黑色边框
    boxrule=1pt,               % 边框线粗细
    boxsep=4pt,                % 内容与边框的内边距（上下左右）
    top=4pt, bottom=4pt,       % 上下边距略增
    left=8pt, right=8pt        % 左右间距扩大
]
\begin{center}
\textbf{Reasoning Template of MiCo}
\end{center}
\textbf{Reasoning Prompt:} First output the thinking process in <think> </think> and give the final answer in <answer> </answer> tags. \\

\textbf{User Question:} Regardless of the augmentation, are image1 and image2 the same? How about image2 and image3, image1 and image3? Only return T(True) or F(False) in <answer> </answer>, for example <think> </think> <answer>TFT</answer>.\\
\end{tcolorbox}

As we verified in Tab.2~(c) that prompt diversity plays a critical role in encouraging the model to generalize its reasoning ability. To this end, we design a systematic prompt expansion strategy along two axes: question phrasing and comparison structure.

First, we construct both forward questions~(\textit{e.g.}, ``Are image1 and image2 the same?'') and reverse questions~(\textit{e.g.}, ``Are image1 and image2 different?''), allowing the model to reason under varied semantic instructions. Second, we vary the combinations of image pairs being queried~(\textit{e.g.}, image1 vs. image2, image2 vs. image3, image1 vs. image3), ensuring comprehensive coverage of possible relations.
Through controlled sampling, we balance the resulting answer types~(\textit{e.g.}, \texttt{TFT}, \texttt{TTF}, \texttt{FFF}, \textit{etc}.), avoiding label bias and promoting robust multi-image understanding across diverse visual scenarios and logical outcomes.

\noindent \textbf{Data filtering.} We apply data filtering When selecting similar but different image pairs from the image editing dataset and video frames. 
For image edting data, to ensure high-quality training samples for visual reasoning, we implement a pixel-wise comparison strategy to filter out image pairs with overly large differences. Specifically, for each image pair, we compute the absolute difference across corresponding pixels. For RGB images, we first average the per-channel differences to obtain a single grayscale difference map. A pixel is considered ``different'' if its value exceeds a predefined threshold~(30). We then calculate the ratio of differing pixels across the entire image. If this difference ratio exceeds 0.8, the image pair is discarded. This simple yet effective rule ensures that the model focuses on learning from subtle and semantically meaningful variations, rather than from trivially dissimilar or noisy pairs. 
For video data, we calculate the SSIM between the selected two frames, and remove the samples with an SSIM value greater than 0.95 to filter nearly the same images.

\noindent \textbf{Differences with NoisyRollout~\cite{noisyrollout}.}
Recently, NoisyRollout~\cite{noisyrollout} also leverages image augmentation to enhance GRPO~\cite{grpo}. However, our Augmented GRPO is fundamentally different from their strategy.

NoisyRollout introduces a hybrid rollout strategy by mixing trajectories from both clean and moderately distorted images. 
Specifically, they add Gaussian noise on the images. Then, sample \text{n} trajectories on noisy images and another \text{n} trajectories on clean images. In this way, they get \text{2n} trajectories in total with more diversity. Afterwards, NoisyRollout calculates the advantages based on the hybrid trajectories and optimizes the policy model on clean images. 

Differently, our Augmented GRPO is inspired by semi-supervised learning in computer vision, where we sample \text{n} trajectories with weak image augmentation. We assume that it would be easier for the policy model to get more high-quality CoTs with weak image augmentation. Then, we calculate the advantages using these high-quality CoTs and optimize the model with stronger augmented images. 
This allows us to obtain correct and informative Chain-of-Thoughts (CoTs) even for samples that would otherwise be answered incorrectly under strong augmentation, thereby improving the model's generalization to more challenging examples.

\begin{table}[t]
\caption{%
    \textbf{Task analysis on MMStar.} We report performance on different sub-tasks to evaluate the generalization ability of \method compared to the Qwen2.5VL baseline.}
\label{tab:mmstar_tasks}
\vspace{-3pt}
\centering\footnotesize
\setlength{\tabcolsep}{4pt}
\scalebox{0.85}{
\renewcommand{\arraystretch}{1.2}
\begin{tabular}{lccccccc}
\hline
& Overall & Coarse Perc. & Fine Perc. & Inst. Reason. & Logic Reason. & Math & Sci. \& Tech. \\
\hline
Qwen2.5VL-7B & 64.07 & 72.00 & 60.40 & 69.60 & 67.20 & 65.60 & 49.60 \\

\method-7B & 
\textbf{65.33} \color{red}{+} & 
\textbf{73.20} \color{red}{+} & 
59.60 \color{green}{-} & 
\textbf{72.00} \color{red}{+} & 
\textbf{68.80} \color{red}{+} & 
\textbf{69.20} \color{red}{+} & 
49.20 \color{green}{-} \\
\hline
\end{tabular}
}
%\vspace{-10pt}
\end{table}

\begin{table}[t]
\caption{%
    \textbf{Task analysis on MuirBench.} Performance breakdown across 12 sub-tasks to evaluate the fine-grained generalization ability of \method.}
\label{tab:muirbench_tasks}
\vspace{-3pt}
\centering\footnotesize
\setlength{\tabcolsep}{4pt}
\scalebox{0.85}{
\renewcommand{\arraystretch}{1.2}
\begin{tabular}{lcccccc}
\hline
& Action & Attr. Sim. & Cartoon & Counting & Diagram & Diff. Spot. \\
\hline
Qwen2.5VL-7B & 40.85 & 58.67 & 46.15 & 34.19 & 77.89 & 54.41 \\
\method-7B & 
40.85 & 
57.65 \color{green}{-} & 
46.15 & 
34.19 & 
\textbf{79.90} \color{red}{+} & 
\textbf{55.29} \color{red}{+} \\
\hline
& Geo. Und. & Img-Text & Ordering & Scene Und. & Vis. Grnd. & Vis. Ret. \\
\hline
Qwen2.5VL-7B & 49.00 & 72.63 & 14.06 & 61.83 & 33.33 & 63.70 \\
\method-7B & 
\textbf{53.00} \color{red}{+} & 
\textbf{74.14} \color{red}{+} & 
\textbf{20.31} \color{red}{+} & 
\textbf{63.98} \color{red}{+} & 
\textbf{35.71} \color{red}{+} & 
\textbf{71.23} \color{red}{+} \\
\hline
\end{tabular}
}
%\vspace{-10pt}
\end{table}

\begin{table}[t!]
\caption{%
    \textbf{Task analysis on BLINK.} Performance comparison across 14 sub-tasks to evaluate the generalization ability of \method.}
\label{tab:blink_tasks}
\vspace{-3pt}
\centering\footnotesize
\setlength{\tabcolsep}{3.5pt}
\scalebox{0.85}{
\renewcommand{\arraystretch}{1.2}
\begin{tabular}{lccccccc}
\hline
& ArtStyle & Counting & Forensic & FuncCorr & IQTest & Jigsaw & MultiView \\
\hline
Qwen2.5VL-7B & 69.23 & 70.83 & 48.48 & 27.69 & 18.00 & 59.33 & 54.89 \\
\method-7B & 
\textbf{72.65} \color{red}{+} & 
70.00 \color{green}{-} & 
47.27 \color{green}{-} & 
\textbf{30.77} \color{red}{+} & 
\textbf{26.00} \color{red}{+} & 
\textbf{69.33} \color{red}{+} & 
42.11 \color{green}{-} \\
\hline
& ObjLoc & RelDepth & RelReflect. & SemCorr & SpatialRel & VisCorr & VisSim \\
\hline
Qwen2.5VL-7B & 54.10 & 81.45 & 40.30 & 33.09 & 88.81 & 52.33 & 86.67 \\
\method-7B & 
\textbf{54.92} \color{red}{+} & 
76.61 \color{green}{-} & 
31.34 \color{green}{-} & 
\textbf{41.73} \color{red}{+} & 
\textbf{90.21} \color{red}{+} & 
\textbf{61.05} \color{red}{+} & 
85.19 \color{green}{-} \\
\hline
\end{tabular}
}
\vspace{-10pt}
\end{table}

\section{More Experimental Results}\label{sec:moreexp}
\subsection{More quantitative results}
To further evaluate the generalization ability of \method, we report its performance on a diverse set of sub-tasks from three comprehensive benchmarks: MMStar~\cite{MMStar}, MuirBench~\cite{muirbench}, and BLINK~\cite{blink}. As shown in Tables~\ref{tab:mmstar_tasks}, \ref{tab:muirbench_tasks}, and \ref{tab:blink_tasks}, \method consistently improves upon Qwen2.5VL-7B across most reasoning-related tasks.

On \textbf{MMStar}, \method achieves notable gains in instance reasoning, logical reasoning, and math, suggesting enhanced multi-step inference and abstraction capabilities. The performance on fine-grained perception slightly decreases, indicating potential room for improvement in precise visual attribute understanding.

For \textbf{MuirBench}, \method improves on 8 out of 12 sub-tasks, including Diagram Understanding, Geographic Understanding, and Visual Retrieval. These tasks involve complex spatial, contextual, or comparative reasoning, showing the effectiveness of our visual comparison objective. Tasks like Attribute Similarity and Counting show marginal drops, possibly due to their reliance on absolute visual matching rather than relational reasoning.

On \textbf{BLINK}, \method shows strong improvements on Functional Correspondence, IQ Test, Jigsaw, Semantic Correspondence, and Visual Correspondence---all of which require visual logic, spatial matching, or multi-view inference. However, tasks such as Multi-view Reasoning and Relative Reflectance exhibit declines, suggesting future efforts could focus on making the model more robust to challenging viewpoint shifts and subtle appearance variations.

Overall, these quantitative results demonstrate that \method is particularly effective at improving tasks involving reasoning, structure, and comparison, while fine-grained low-level perception remains a direction for future enhancement.

\subsection{Unsuccessful Attempts}
Throughout our exploration, we experimented with several alternative approaches that ultimately did not lead to improved performance. For completeness and to facilitate future research, we summarize these unsuccessful attempts and provide insights into why they may have failed.

\noindent \textbf{Confidence reweighting.} Since our task is formulated as answering T/F questions, even when evaluating three comparisons simultaneously, there remains a non-trivial chance~(12.5\%) of obtaining the correct answer purely by guessing. To reduce the impact of such randomness, we explored adding an additional reward or weight based on the model's answer confidence. Specifically, we experimented with several approaches to compute confidence scores from the softmax probabilities of the output tokens. However, these confidence-based reweighting strategies did not yield any performance improvements. We analyze that this may be due to the fact that the softmax probability of the predicted token does not reliably reflect the model's true certainty about the overall answer. In particular, the model may assign high confidence to tokens that are syntactically or semantically unrelated to the actual correctness of the reasoning (e.g., punctuation, or irrelevant words within the output). As a result, the computed “confidence” can be misleading, making it an ineffective signal for reward shaping.

\noindent \textbf{Importance sampling.} As in our Augmented GRPO, we sample the trajectories on simple examples with weak augmentations, but we use the trajectory to optimize harder exaples with strong augmentations. This might cause misalignment similar to offline reinforcement learning. In this way, we apply importance sampling, which calculates the probability gap between the trajectories for the simple and hard examples as a weight to reweight the reward/advantages. This strategy could not bring improvements. 
We suspect that although importance sampling is theoretically justified, it may interfere with the core optimization dynamics of GRPO. Specifically, GRPO relies on the relative ranking of trajectories within a group to compute structured advantages. Introducing importance weights—derived from distribution shifts—may distort this internal ranking or inject instability into the reward signals. Additionally, the token-level probability changes caused by visual augmentations can be noisy or poorly calibrated, making the computed importance weights unreliable in practice.

\section{Qualitative Analysis}\label{sec:morequa}
\begin{figure}[t]
\centering 
\includegraphics[width=0.9\linewidth]{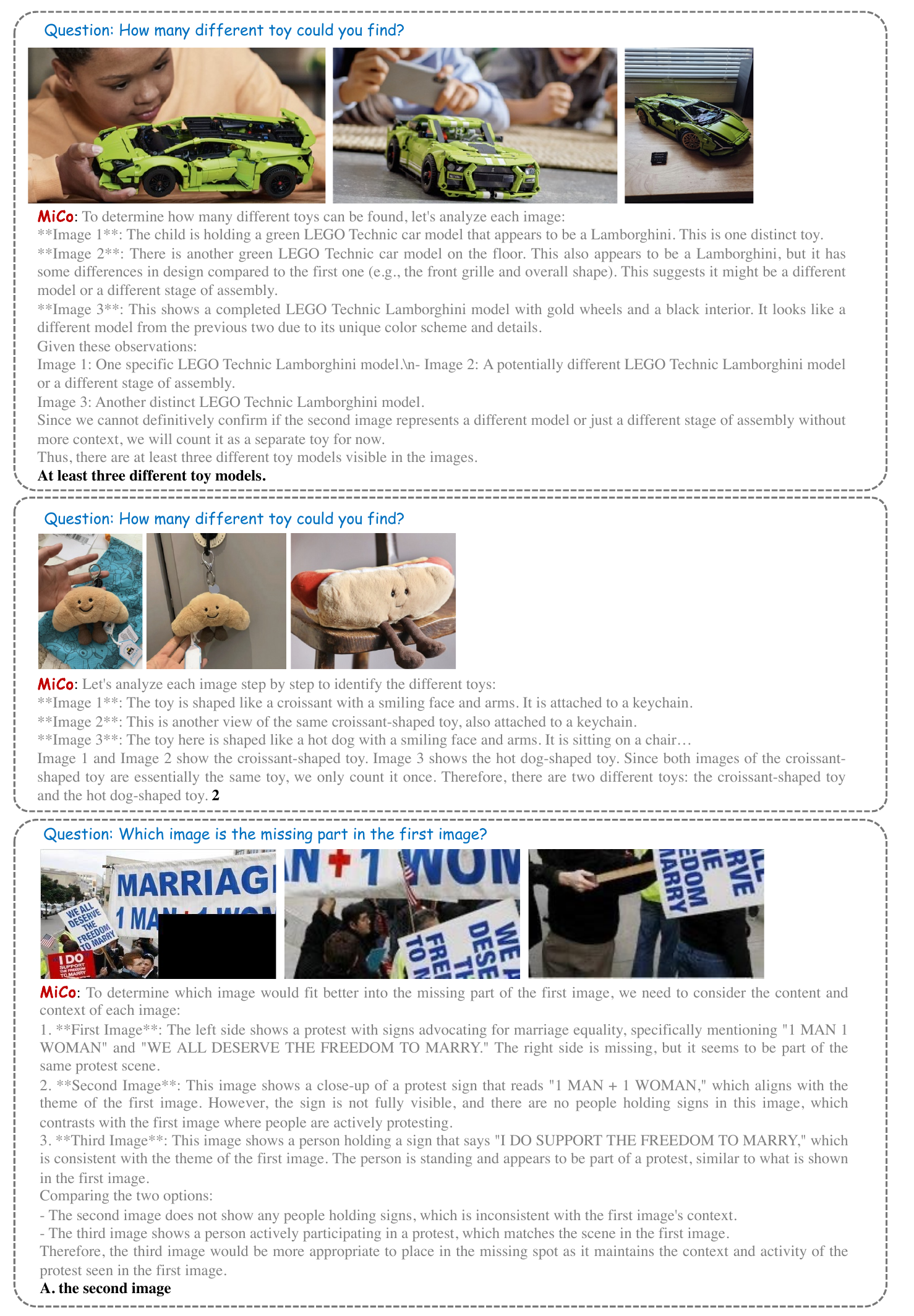} 
\vspace{-8pt}
\caption{%
    \textbf{Demonstrations} for detailed comparison and jigsaw solving.} 
\label{fig:demo1}
\vspace{-8pt}
\end{figure}

\begin{figure}[t]
\centering 
\includegraphics[width=0.9\linewidth]{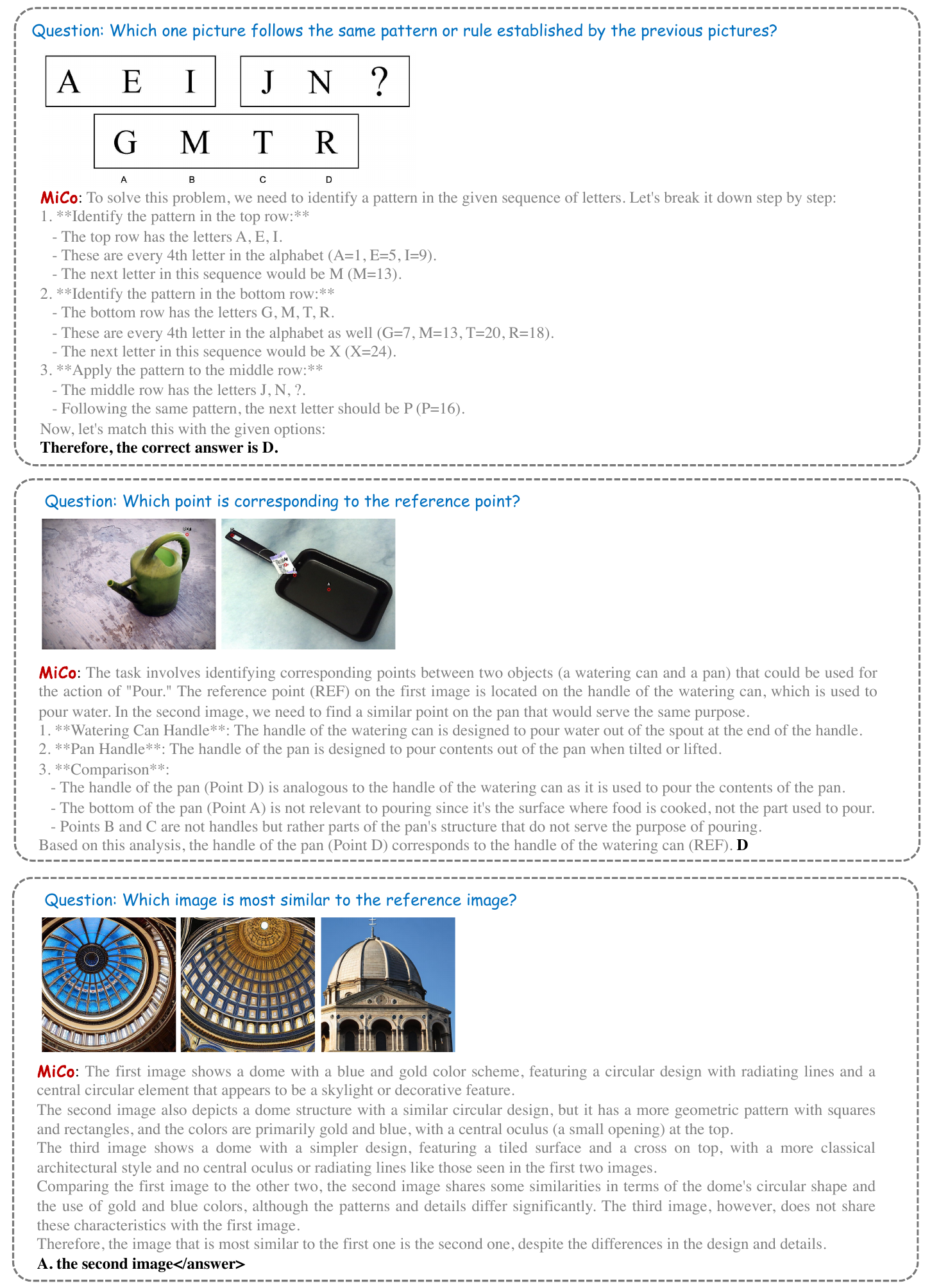} 
\vspace{-8pt}
\caption{%
    \textbf{Demonstrations} for IQ test, functional correspondence, and visual similarity.} 
\label{fig:demo2}
\vspace{-8pt}
\end{figure}

We add more visual demonstrations for the reasoning ability of \method in \cref{fig:demo1} and  \cref{fig:demo2}. 
These qualitative examples demonstrate the strong reasoning capability of \method across various visual tasks. In Figure~\ref{fig:demo1}, the model exhibits detailed step-by-step analysis to distinguish visual differences, count distinct objects, and solve jigsaw-like problems. Rather than relying on superficial features, \method actively grounds its reasoning in object identity, pose, structure, and scene semantics. For example, in the toy comparison case, it not only detects the number of different objects but also considers subtle variations in assembly, model type, and color configuration. In the jigsaw task, it correctly identifies missing or manipulated segments by referencing spatial consistency and scene-level context.

Figure~\ref{fig:demo2} further highlights \method’s ability to tackle abstract reasoning challenges. In the IQ-style pattern recognition question, the model deduces a complex symbol progression rule based on character groupings and positions. For functional correspondence and spatial matching, it accurately aligns image pairs by understanding object affordances and relative part placement. Additionally, in the visual similarity task, it discerns fine-grained geometric and design attributes to match images at a structural level rather than based on superficial texture or color.

Together, these examples reveal that \method does not merely perform image-text matching but is capable of systematic, multi-step reasoning grounded in visual understanding. This reflects its generalization ability across both low-level visual tasks and high-level abstract reasoning challenges.

\section{Potential Social Impact}\label{sec:impact}
MiCo explores a self-supervised and reinforcement learning-based approach to improve multi-image reasoning in vision-language models without relying on human-annotated question-answer pairs. By leveraging intrinsic visual constraints, such as consistency across augmented views and differences between similar images, MiCo significantly reduces the need for labor-intensive data curation. This has the potential to democratize the development of reasoning-capable AI systems, making them more accessible in low-resource settings or for underrepresented languages and domains where curated datasets are scarce.

However, as with any powerful vision-language technology, there is a risk of misuse, particularly in applications involving surveillance, misinformation, or unauthorized inference of user intent from visual data. MiCo's improved ability to perform fine-grained comparisons across images could be exploited in privacy-invading scenarios if deployed irresponsibly. To mitigate such risks, we advocate for deploying MiCo in alignment with responsible AI guidelines, ensuring transparency, consent, and clear boundaries in its application domains. In practice, this includes integrating robust sensitive content filtering, restricting deployment in high-stakes or privacy-sensitive scenarios, and establishing human-in-the-loop mechanisms for critical decision-making processes.

\end{document}